%% file: main.tex
\documentclass[10pt,twocolumn,letterpaper]{article}

\usepackage[pagenumbers]{wacv} 

\usepackage{graphicx}
\usepackage{amsmath}
\usepackage{amssymb}
\usepackage{booktabs}

\usepackage[pagebackref,breaklinks,colorlinks]{hyperref}

\usepackage[capitalize]{cleveref}
\usepackage{xcolor}

\crefname{section}{Sec.}{Secs.}
\Crefname{section}{Section}{Sections}
\Crefname{table}{Table}{Tables}
\crefname{table}{Tab.}{Tabs.}

\def\wacvPaperID{102} 
\def\confName{WACV}
\def\confYear{2024}
\def\methodName{PRISM}

\newcommand{\TODO}[1]{\textcolor{red}{TODO:#1}}

\begin{document}
\newcommand{\pavel}[1]{\textcolor{orange}{#1}}

\title{PRISM: Progressive Restoration for Scene Graph-based Image Manipulation}

\author{Pavel Jahoda$^1$, Azade Farshad$^1$, Yousef Yeganeh$^1$, Ehsan Adeli$^2$, and Nassir Navab$^{1,3}$\\
$^1${\normalsize Technical University of Munich}\\
$^2${\normalsize Stanford University}\\
$^3${\normalsize Johns Hopkins University}
}
\maketitle

\input{chapters/0_abstract}

\input{chapters/1_intro}
\input{chapters/2_relworks}
\input{chapters/3_method}
\input{chapters/4_results}
\input{chapters/5_conclusion}

{\small
\bibliographystyle{ieee_fullname}
\bibliography{bibliography}
}

\end{document}

%% file: chapters/0_abstract.tex
\begin{abstract}
Scene graphs have emerged as accurate descriptive priors for image generation and manipulation tasks, however, their complexity and diversity of the shapes and relations of objects in data make it challenging to incorporate them into the models and generate high-quality results. To address these challenges, we propose PRISM, a novel progressive multi-head image manipulation approach to improve the accuracy and quality of the manipulated regions in the scene. Our image manipulation framework is trained using an end-to-end denoising masked reconstruction proxy task, where the masked regions are progressivley unmasked from the outer regions to the inner part. We take advantage of the outer part of the masked area as they have a direct correlation with the context of the scene. 
Moreover, our multi-head architecture simultaneously generates detailed object-specific regions in addition to the entire image to produce higher-quality images. Our model outperforms the state-of-the-art methods in the semantic image manipulation task on the CLEVR and Visual Genome datasets. Our results demonstrate the potential of our approach for enhancing the quality and precision of scene graph-based image manipulation.

\end{abstract}

%% file: chapters/1_intro.tex
\vspace{-10pt}
\section{Introduction}
Image generation and manipulation have seen significant advancements in recent years, with autoencoders \cite{VAE} and GANs \cite{GAN} being the primary approaches in the early years. Generator architectures have progressed substantially in complexity, with the development of architectures such as SPADE \cite{SPADE_2019} and Pix2Pix \cite{pix2pixHD_2018}. While recent research on diffusion models \cite{diffusion_original} has gained interest, GANs remain a popular choice due to their ability to achieve competitive performance in generating high-quality images.

\begin{figure}[t!]
    \centering
    \includegraphics[width=\linewidth]{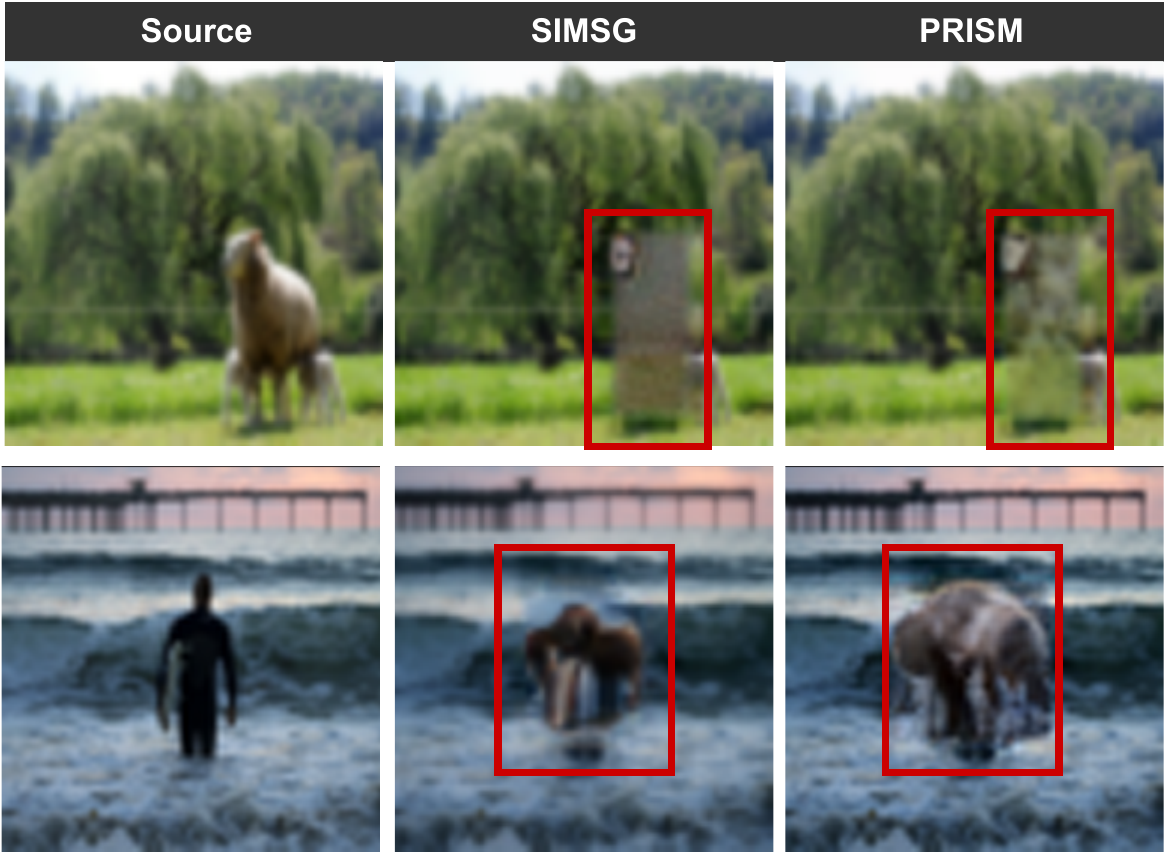}
    \caption{Our method demonstrates significant improvement in capturing image details compared to SIMSG~\cite{simsg} as shown through two different image manipulaton tasks: object removal \textbf{(top row)}, replacing a \underline{man} with an \underline{elephant} \textbf{(bottom row). }}
    \label{fig:teaser}
\end{figure}


Conditional image generation \cite{CGAN}, particularly on different modalities such as text \cite{dalle2_ramesh2022} and image \cite{pix2pixHD_2018}, has been an interesting topic in research. Scene graph-based \cite{johnson2015image} image generation \cite{sg2im} has emerged as a promising direction in conditional image generation, enabling the generation of complex images from natural and intuitive language input. In addition, image manipulation approaches alter images based on input priors such as segmentation maps \cite{SESAME}, text \cite{li2020manigan} or scene graphs \cite{simsg}. Semantic image manipulation methods such as SIMSG \cite{simsg} provide a means of object manipulation as well as changes in the structure of the scene. For example, they are able to change the relationships between the objects without requiring pairs of annotated source and target data.
However, these methods face many challenges that reduce their effectiveness. A common challenge is the lack of context information of the data, which hinders the learning of complete representation of the scene and objects relationships. Another challenge is the complexity and unabalanced distribution of the scenes in datasets such as visual genome, reducing the performance of the model. A third challenge is the uncertainty of some relationships such as \textit{near} or \textit{by}, leading to ambiguity for image manipulation. These factors result in low image quality, limiting the practical applications of scene graph-based image generation and manipulation techniques.

To address these challenges, we propose PRISM, a new progressive multi-head semantic image manipulation approach. Our method leverages image reconstruction through a progressive denoising process for image manipulation using scene graphs. This scene graph-based approach does not require direct supervision for constellation changes or image edits and provides additional context information to enable more precise image manipulation.  Our multi-head architecture simultaneously reconstructs the entire image and selected image objects in detail, resulting in high-quality and detailed images that are more faithful to the original scene graph. 

In summary, our contributions are:  First, we present a novel multi-head approach for semantic image manipulation. The multi-head approach is realized through a GAN featuring a decoder comprised of two heads. One head focuses on generating detailed generation of the manipulated objects while the other focuses on holistic image manipulation. Second, we propose a progressive denoising approach for the image reconstruction proxy task, which provides additional context information enableing more precise image editing. We evaluate these techniques quantitatively in an image reconstruction task since there are no pairs of data for before and after the changes and qualitatively for image manipulation. Finally, we combine these techniques into an end-to-end progressive multi-head image manipulation approach that outperforms the previous work on semantic image manipulation.

%% file: chapters/2_relworks.tex
\section{Related Work}
Recent advances in computer vision and deep learning have led to significant progress in the field of image generation and manipulation from GANs\cite{GAN} to diffusion\cite{diffusion_original} models. Despite the high-quality results of diffusion models, GANs can be very problem-specific, and they are often customized or fine-tuned to serve a particular purpose, hence, we have varieties of GAN-based models: DCGAN\cite{DCGAN} improved the quality images utilizing convolutional layers. InfoGAN\cite{chen2016infogan} aims to learn disentangled information by giving semantic meaning to features in the latent space. CoGAN\cite{liu2016coupled} uses a pair of GANs to learn the joint distribution of multi-domain images with weight-sharing constraints, which requires fewer parameters and resources than two individual GANs; later, AC-GAN\cite{shu2017ac} and StackGAN\cite{zhang2017stackgan} were introduced for improved synthetic image generation. WGAN\cite{arjovsky2017wasserstein} modified the training phase to update the discriminator more often than the generator avoiding mode collapse. CycleGAN\cite{CycleGAN2017} use an image-image generation without paired data, consisting of two generators with the addition of cycle consistency loss to ensure image similarity. SAGAN\cite{zhang2019self} maintains long-range relationships within an image and uses spectral normalization. BigGAN\cite{brock2018large} produces high-quality images by upscaling existing GAN models and introducing techniques to detect training instability.

\paragraph{Conditional Image Generation}
CGAN \cite{CGAN} and AC-GAN \cite{shu2017ac} allow generated data to be conditioned on class labels, enabling the generation of synthetic samples based on specific class labels in labeled datasets. StyleGAN \cite{karras2019style} adds more control over generated images through separating high-level features and stochastic variation, rather than solely generating realistic images, and GauGAN \cite{park2019semantic} employs a spatially-adaptive normalization layer to generate photorealistic images from user-provided abstract scene sketches while allowing scene manipulation and element labeling. Object layouts are used as a prior in \cite{sylvain2021object, li2021image} to deal with overlapping objects and \cite{he2021context} to address lack of context aware modeling. Recently, textual conditioning of image generation models gained attention. CLIP (Contrastive Language-Image Pre-Training) \cite{CLIP_2021} and word2vec \cite{word2vec} are commonly used language models for generating images based on textual inputs, through transfer learning and efficient word representation estimation, respectively. Utilizing that diffusion models like DALL-E \cite{dalle_ramesh2021}, DALL-E2 \cite{dalle2_ramesh2022}, GLIDE \cite{diffusion_2021glide} and Stable-Diffusion \cite{rombach2022high}, are able to generate realistic images. Semantic Layout to Image Generation with Transformers \cite{semantic_layout} utilizes hierarchical text that can be transformed into images.

\paragraph{Progressive Image Generation}

In recent years, progressive image generation techniques have been widely explored in computer vision and have shown significant improvement over traditional image generation methods. Gao et al. \cite{gao2019progan} proposed a method for training GANs by gradually increasing the resolution of generated images during training. This stabilizes and accelerates GAN training, resulting in high-quality generated images. In interactive and data-driven image editing, Hong et al. \cite{sem_img_manipulation} enabled manipulation of images using object bounding boxes with an adaptive determination of structure and style. The Progressive Growing of GANs (PGN) \cite{progressive_handcrafted} is another end-to-end framework for semantic image inpainting, using curriculum learning by dividing the hole filling process into several different phases. FRRN~\cite{progressive_learned_2019} and FRF~\cite{rfr_li2020recurrent} fill irregular holes with a step loss function and N Blocks. FRPN implements One Dilation strategy, while FRF incorporates attention mechanism to improve the quality of intermediate restorations.


Dhamo et al. \cite{inpainting_dhamo_2019} introduced a pipeline involving foreground removal, GAN-based inpainting for missing regions, and a pair discriminator for inter-domain consistency. S2-GAN \cite{style_structure_wang_2016} proposed a method for learning using independent learning of Style-GAN and Structure-GAN and joint end-to-end learning. SESAME \cite{SESAME} is an adversarial learning-based semantic editing architecture that provides full control over image output by pixel-level guidance of semantics. These works demonstrated the effectiveness of progressive image generation in various tasks such as dynamic image enhancement \cite{progressive_handcrafted}, image compression \cite{progressive_learned_2019}, and image inpainting \cite{inpainting_dhamo_2019}. Moreover, Laplacian Pyramid as a Compact Image Code \cite{LDI_1998} also demonstrated the effectiveness of progressive image generation.
\paragraph{Image Manipulation}
Image manipulation is an image generation task that keep some aspects of an image fixed and generating the rest based on a prior or condition.  SPADE~\cite{SPADE_2019} uses input semantic layout to perform affine transformation. Semantic Explicitly-Supervised GAN for Cross-Modal Retrieval~\cite{SESAME} is another work that manipulate images with pixel-level guidance of semantic labels for only the regions to be edited. SIMSG~\cite{simsg} enabled flexible changes and additions while preserving the original content and respecting semantics and style.

\paragraph{Image Generation and Manipulation using Scene Graphs}
Image generation and manipulation using scene graphs has become a popular area of research in recent years. Scene graphs are powerful tools for accurately manipulating images by representing the objects and their relationships to one another in a scene, with each object as a node. Researchers have proposed various methods and approaches for generating and manipulating images using scene graphs.

Some methods have focused on generating images from scene graphs using graph auto-regressive models, such as the Maximum Mean Discrepancy (MMD) metric with random-walk graph and node kernels, as proposed by Garg et al.~\cite{garg2021unconditional}. Other methods, such as the Meta-learning-based Image Generation System (MIGS)~\cite{azade_2021_MIGS}, utilize a meta-learning approach for generating images from scene graphs. Disentangling pose and identity in image manipulation has also been explored. DisPositioNet, proposed by Farshad et al.~\cite{farshad2022DisPositioNet}, uses disentangled feature extracted from the scene graphs to disentangle pose and identity in image manipulation.

Some methods have focused on generating images that encode relationships using relation affinity fields. For example, Factorizable Net~\cite{factorizable_net_2018} incorporates bottom-up clustering, spatial weighted message passing, and spatial-sensitive relation inference modules while maintaining spatial information and improving relationship recognition between objects. Other methods have addressed the problem of logical equivalence in generating layouts from scene graphs. Herzig et al.~\cite{scene_graph_inverse} addressed this issue by using canonical representations of scene graphs, resulting in stronger invariance and generalization properties.

Several approaches have also been proposed for generating images from scene graphs by introducing scene context networks, context-aware loss, and defining new evaluation metrics. For example, Tripathi et al.~\cite{scene_graph_context_image_gen_2019} proposed an approach to generate images from scene graphs by using a scene context network and context-aware loss, and defining new evaluation metrics such as the relation score and mean opinion relation score (MORS) to measure compliance of the generated images with the scene graph.

Other methods have utilized temporal dependencies across frames to decode visual relationship representations effectively, such as STTran~\cite{scene_graphs_transformers} that generates a dynamic scene graph from sequences by applying multi-label classification in relationship prediction. Finally, some methods have utilized external object crops as anchors to generate realistic images from a scene graph, such as PasteGAN~\cite{scene_graph_image_gen_2019pastegan} or RetrieveGAN~\cite{scene_graph_image_gen_2020retrievegan}. Meanwhile, Xu et al.~\cite{scene_graph_jointly_2017Xu} presented an end-to-end model for generating visually-grounded scene graphs from images by iteratively refining its prediction through contextual messaging along the topological structure of the scene graph.

%% file: chapters/3_method.tex
\section{Methodology}\label{sec:method}

\begin{figure*}[t]
    \centering
    \includegraphics[width=\textwidth]{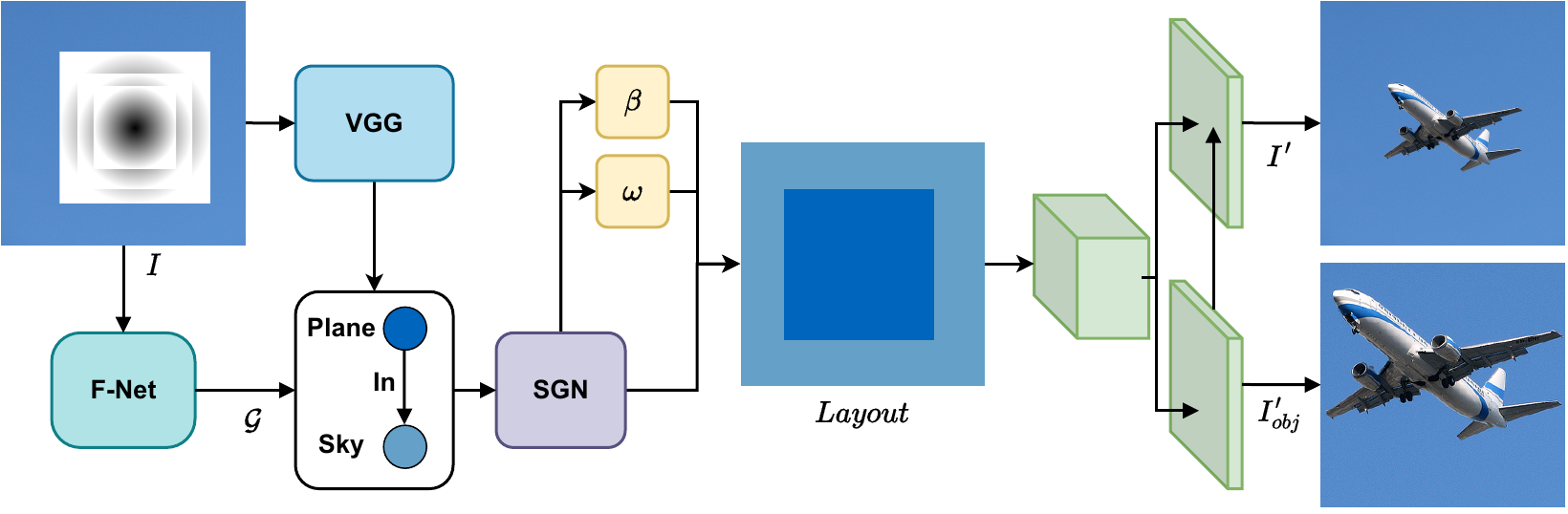}
    \caption{\textbf{ \methodName{} Overview.} a) During self-supervised training through reconstruction, the model first predicts a scene graph from the image, then it randomly masks an object in the scene, and predicts a scene layout from the image using the Scene Graph Network. Finally, the image is progressively generated through the decoder by filling the border regions of the masked input. The model progressively reconstructs the entire original image in the first head (top-right) and reconstructs the masked object region in the second head in detail (bottom-right). b) At test time, the image manipulation is performed on a source images by manually modifying the scene graph to obtain the target image.}
    \label{fig:method}
\end{figure*}



In this section, we give a detailed description of our methodology. An overview of our methodology is depicted in~\autoref{fig:method}.


\paragraph{Definitions}
We are given a dataset $\mathbb{D} = (I, \mathcal{G})$ of input images $I \in \mathbb{R}^{W \times H \times 3}$, and their corresponding scene graphs $\mathcal{G} = (V,E)$ where $V$ and $E$ are the vertices and edges in the scene graphs. The network is denoted by $\theta$, where it has two outputs $I', I'_{obj} = \theta(I, \mathcal{G})$. The scene graph $\mathcal{G}$ is predicted using a pre-trained scene graph generation model. Each scene graph $\mathcal{G}_i$ consists of a set of triplets $(o_{ij}, p_{ij}, s_{ij})$ where $o, p, s$ are the object, predicate and subject and $j \in J$, where $J$ is the total number of triplets in the given scene graph.

In the training process, the model is trained through a reconstruction objective where the input image $I$ and scene graph nodes are randomly masked and the masked area is reconstructed by the model from the extracted scene graph features. The network is trained in a multi-head setting, where a randomly masked object is generated by a second decoder head in addition to the whole image generated by the main decoder. The reconstructed image and object are denoted by $I', I'_{obj}$ respectively. 

\subsection{PRISM Architecture}
\paragraph{Image to Scene Graph}\label{sec:im_to_sg}
In the first part of the training/inference pipeline, we first preprocess the images $I$ and then feed them into a pre-trained scene graph generation model. Here we take Factorizable Net~\cite{factorizable_net_2018} (F-Net) as the state-of-the-art model on the visual genome dataset. F-Net uses a Region Proposal Network~\cite{faster_rcnn} to generate object region proposals. It then uses a bottom-up approach where these region proposals are clustered into subgraph scene predictions. Messages are then passed between subgraphs to maintain spatial information and to refine the feature representation of the subgraphs. Finally, objects are predicted from the object features, and predicates are inferred based on the object features and the subgraph features. 

\paragraph{Scene Graph Representation}
Scene graphs are a structured representation of a scene consisting of a set of nodes typically corresponding to objects and a set of edges that describe the relationship between the objects. We form the node features by enriching the nodes with semantic information. The node feature vector representation consists of three parts which are concatenated together: (1) the object class embeddings from a learnable embedding layer, (2) the object's bounding box, and (3) a visual feature encoding of the object obtained from the pre-trained VGG \cite{vgg} network.

The node features are processed using a graph convolutional network called Spatio-semantic Scene Graph Network (SGN) that allows feature representation information to flow through the graph. The edges are updated as given in \autoref{eq:edge_update}:

\begin{equation}\label{eq:edge_update}
(\gamma_{ij}^{(t+1)}, \rho_{ij}^{(t+1)}, \zeta_{ij}^{(t+1)}) = \tau_{edges}(v_{i}^{(t)},
\rho_{ij}^{(t)},v_{j}^{(t)})
\end{equation}

where $v_{i}^0$ corresponds to the $i$-th node feature vector before scene graph processing, $\tau_{edges}$ transformation is a implemented as a multilayer perceptron, and $\gamma$, $\rho$, and $\zeta$ correspond to latent vector representation of object, predicate, and subject respectively. The node update depicted in \autoref{eq:node_update} is computed as an average of the edge update result:

\begin{equation}\label{eq:node_update}
v_{j}^{(t+1)} = \tau_{nodes}(\dfrac{1}{N_{i}}(\sum_{j|(i,j)\in \mathbb{R}}^{} \gamma_{ij}^{(t+1)} + \sum_{k|(k,i)\in \mathbb{R}}^{} \zeta_{ki}^{(t+1)})   )
\end{equation}

where $N_{i}$ is the number of neighbors of the $i$-th node and transformation $\tau_{nodes}$ is implemented as a multilayer perceptron. We obtain a set of object features $v_{j}$, from which we predict the bounding box and pseudo-segmentation map for each object. We denote the bounding box prediction and pseudo-segmentation map prediction networks which are multi-layer perceptrons by $\beta$ and $\omega$ respectively. The SGN learns a robust object representation which encodes spatial relationship information and attribute binding for objects.

\paragraph{Layout Generation}
The processed scene graph representation is then used to produce a 2D scene representation called scene layout. The scene layout is a mapping of object features $v_j$ to their corresponding predicted bounding box locations multiplied by the pseudo-segmentation map predictions.
This 2D spatial arrangement of features is then appended by low-level visual features of the query image obtained by passing the query image through a convolutional neural network. 

\paragraph{Multi-head Image Generation}\label{sec:two_headed_method}
To generate the final image, we feed the constructed layout to the decoder network. The reconstructed image $I'$ and object patch $I'_{obj}$ are generated by the decoder, which is based on the SPADE~\cite{SPADE_2019} architecture.
To enforce the generation of object details in the network, we modify the decoder network to have two heads.
The SPADE decoder consists of multiple residual spade blocks which focus on the normalization in the image generation process. In contrast to batch normalization~\cite{batch_normalization}, which is unconditional and isn't spatially sensitive, SPADE conditions the normalization on the specified input. In batch normalization, there are several learned affine layers that are applied after the normalization step. In SPADE, these layers are learned as the output of convolution on the semantic segmentation map. Such a conditional approach makes the affine layer spatially adaptive. 
 In our model, the generation is conditioned on the masked RGB input images as well as the predicted scene layout (only layout is used in the SPADE normalization).


With the first head, the decoder tries to reconstruct the original entire image. In the second head, the decoder focuses on reconstructing only a single image region containing a previously masked object. The region is defined by a window size of $\delta \times \delta$, where $\delta = w$ if $w>=h$ otherwise $\delta = h$, where $h$ and $w$ represent the height and width of the object to be reconstructed respectively. Generating the magnified object allows the network to gain additional fine detail information about the objects it tries to reconstruct. We also propose a residual connection from the object head to the main head to facilitate the flow of information between the two heads.


\paragraph{Progressive Generation}\label{sec:progressive_method}
Since the image manipulation model is trained through the image reconstruction proxy task, the model is dependent on random masking of parts of the image. Therefore, we propose a progressive unmasking approach for the reconstruction goal.

Our model consists of an encoder and a decoder. The encoder generates an intermediate layout representation from the scene graph. The decoder takes the layout representation and a masked image feature representation as input and generates the output image. The masked image feature representation is obtained by masking image regions given predefind bounding box regions and then feeding it into a shallow one-layer convolutional neural network that produces the low-level feature representation. 
We create a feature representation of the original image by introducing sampled Gaussian noise to the object regions we want to reconstruct. The noise is sampled from a normal distribution with zero mean and standard deviation $\sigma$, i.e., $\mathcal{N}(0, \sigma^2)$. This noisy image and the layout representation are the input to the decoder which generates an output image by reconstructing the masked regions.

In the first iteration, we feed the fully masked image as input to the convoulational network, and then let the decoder generate image $I'$. In the second iteration, we replace a border region of the masked image with pixels from the generated image from the first iteration:
\begin{align*}
        I_{k+1}[0.75w:w,0.75h:h] &=I'_{k}[0.75w:w,0.75h:h] \\ I_{k+1}[0:0.75w,0:0.75h] &\sim \mathcal{N}(0, \sigma^2) 
\end{align*}


\paragraph{Progressive Multi-head Generation}
In the full version, we apply the proposed progressinve denoising approach to the multi-head architecture. In the first head, which is responsible for reconstructing the entire image, we progressivly mask the object regions as described in the previous paragraph. For the second head, which reconstructs only a single object region, the outer region of the object ($[0.75w:w, 0.75h:h]$) is reconstructed in the first pass, and the entire region ($[0:w, 0:h]$) is reconstructed in the second pass.


\subsection{Training}
There are two line approaches with respect to the training data used for training semantic image manipulation models. The first approach, which is resource costly involves having an image pair of an image before modification and an image after modification. The model is simply fed an image before modification and attempts to generate the desired modification according to the ground-truth after-modification image.
In our work, we do not require such image pairs. Instead, we train the model using a single image and performing image reconstruction. During training, we mask the visual feature encoding part of an object's node representation with a probability $p_{\rho}$. Afterward, we mask corresponding object regions in the input image with Gaussian noise. The masked representation is subsequently fed into a network that generates a realistic image resembling the original unmasked image. Additionally, we independently mask the object's bounding boxes with probability $p_{b}$. As a result, the SGN performs a form of image reconstruction as it attempts to predict the bounding boxes of each object. Therefore the model performs two types of reconstruction -- visual and spatial. 

\paragraph{Losses}
We employ common losses in generative adversarial networks for the training. For the generator, we use mean absolute error (L1 loss) between the generated and ground-truth images, mean squared error (L2 loss) between predicted bounding boxes of objects and ground-truth, Perceptual Loss~\cite{perceptual_johnson2016}, Feature Matching Loss~\cite{improved_GANS_2016,pix2pixHD_2018}, and Hinge Loss the objective main objective function of the generative adversarial network. Finally, we use an auxiliary classification loss~\cite{auxillary_loss_2017} that performs a classification of the individual reconstructed objects. This is all trained using multi-scale discriminator~\cite{multiscale_2018high}. We've also experimented with Total Variational Loss~\cite{tv_loss_rudin1992} that is used to encourage spatial smoothness in the generated images. 

\begin{equation}
    \mathcal{L}_{rec} = \| I' - I \|
\end{equation}
\begin{equation}
    \mathcal{L}_{bbox} = \| \beta(I) - b \|_{2}
\end{equation}
, where $b$ is the ground truth bounding box information.
\begin{equation}
  \begin{split}
    \mathcal{L}_{total} = \mathcal{L}_{GAN,img} + \mathcal{L}_{GAN,obj} + \mathcal{L}_{bbox}  \\ + \mathcal{L}_{rec} + \mathcal{L}_{aux, obj}
      \end{split}
\end{equation}

\subsection{Inference}
During the inference phase, the model receives a source image. The scene graph for this image is predicted using the F-Net. Then the user specifies the manipulation by directly modifying the nodes and edges of the predicted scene graph and the model outputs the modified image. The modification is done through masking either the visual features or the bounding box information based on the modification mode. For instance, when an object is removed the box information is retained, while the visual features are masked to be reconstructed. On the other hand, for relationship changes, the box information is masked and predicted by the model, while the visual features are kept.

%% file: chapters/4_results.tex
\section{Experiments}\label{sec:experiments}
\begin{figure*}[t]
    \centering
    \includegraphics[width=\textwidth]{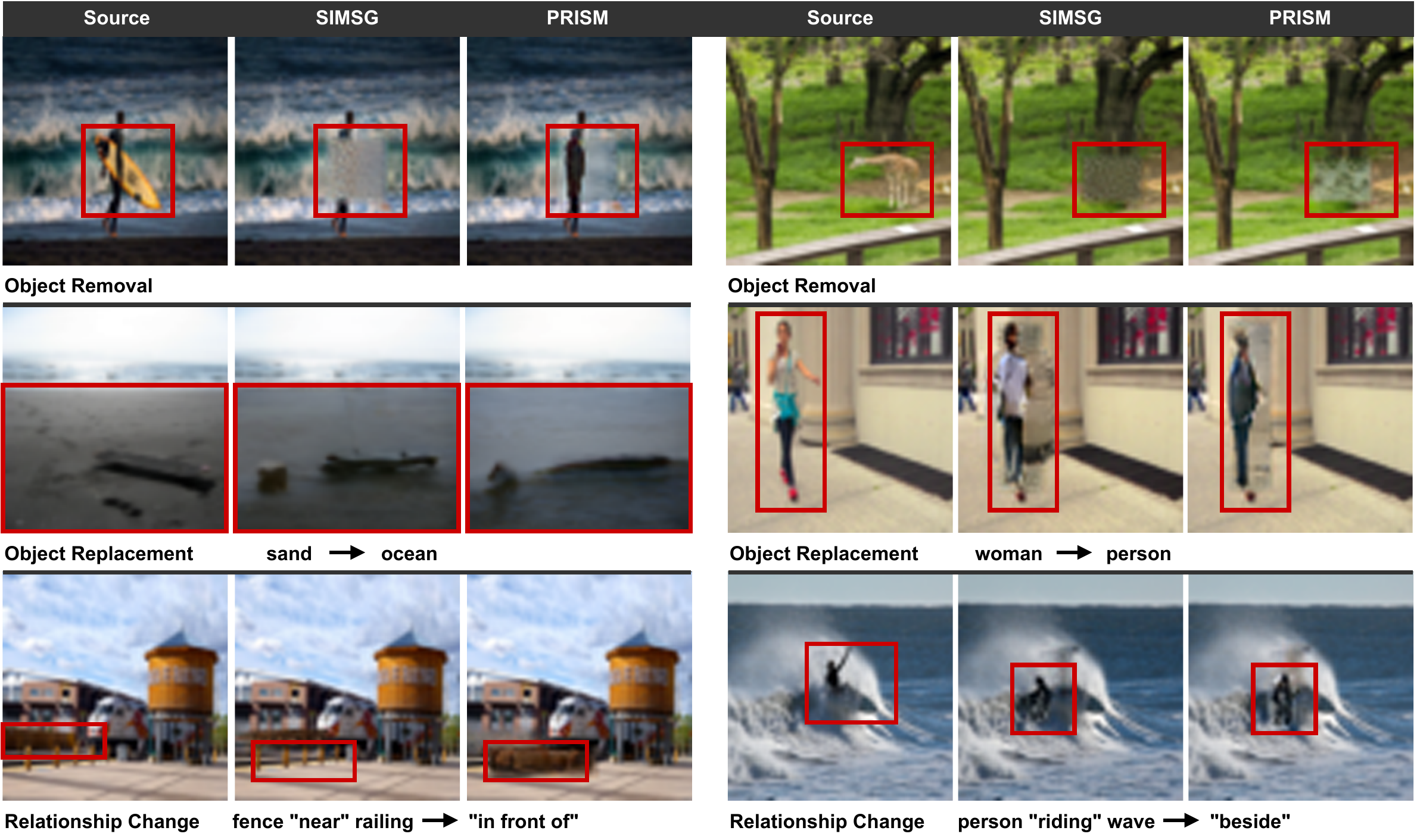}
    \caption[Image manipulation comparison between SIMSG and PRISM]{Qualitative comparison between SIMSG~\cite{simsg} and PRISM in terms of object removal, replacement, and relationship change on the VG \cite{VG_2017} dataset. Or model achieves better image manipulation performance compared to SIMSG in different manipulation modes.}
    \label{fig:comparison_PRISM_SIMSG_VG}%
\end{figure*}

In this section, we provide the results of our experiments on two common benchmarks for semantic image manipulation using scene graphs.
\subsection{Experimental Setup}
\paragraph{Data}
We evaluated our method on two common benchmarks for visual understanding and reasoning, namely CLEVR~\cite{CLEVR} and Visual Genome~\cite{VG_2017}. As opposed to VG, which contains real-life images, CLEVR is a synthetic dataset with a limited complexity of the depicted scenes. Both datasets provided scene graph annotation for each image, making these datasets suitable for the semantic image manipulation task. We follow the pre-processing steps of SIMSG~\cite{simsg} on both datasets. We normalize the mean and variance of the images and resize them to $64 \times 64$ resolution. 

\paragraph{Hyperparameters}
The batch size for training the model was set to $32$. We used a loss weight of $1$ for the L1 pixel loss and the Hinge Loss, a weight of $5$ for the Perceptual Loss~\cite{perceptual_johnson2016} and Feature Matching Loss~\cite{improved_GANS_2016}, a weight of $50$ for the bounding box L2 Loss, and a weight of $0.1$ for the auxiliary classification loss~\cite{auxillary_loss_2017}. 



\paragraph{Training Hardware}
All training was done on an NVIDIA Tesla V100 GPU with $16$ GB of memory. The models require approximately 4 to 6 days for $100K$ optimization iterations of training for CLEVR and 6 to 9 days for $150K$ iterations on the VG dataset.

\paragraph{Evaluation Metrics}
We evaluated our models with multiple metrics used for the assessment of image generation models. Since the quantitative evaluation of image manipulation is not possible in real-world datasets such as the VG dataset, we follow the previous work and measure the image reconstruction performance using the common reconstruction metrics in addition to the Fréchet Inception Distance~\cite{FID_2017} (FID). For the reconstruction, we measure the Mean Absolute Error (MAE), Structural Similarity Index Measure~\cite{SSIM_2004} (SSIM), Learned Perceptual Image Patch Similarity~\cite{LPISP} (LPIPS) between the generated and ground-truth images. 

\subsection{Quantitative Results}\label{sec:quantitative_results}
\paragraph{Comparison to Previous Work}
As we can see in \autoref{tab:main_CLEVR}, both the two-headed approach and the progressive approach significantly outperform the baseline SIMSG~\cite{simsg} architecture in all metrics on the CLEVR dataset. They outperform the baseline approach by a significant margin in both metrics that capture pixel differences (MAE) as well as in metrics that capture high-level image differences (LPIPS~\cite{LPISP}, FID~\cite{FID_2017}).

\begin{table}[tbh]
\centering
\caption[Image reconstruction comparison between main approaches on VG]{Image reconstruction results on the VG compared to the SOTA}
\resizebox{\linewidth}{!}{
\begin{tabular}{l|lllll}
\hline
                      & MAE $\downarrow$      & SSIM $\uparrow$ & LPIPS $\downarrow$ & FID $\downarrow$ & IS $\uparrow$ \\ \hline
ISG~\cite{scene_graph_interactive_image_gen_2019}  & $46.44$ & $0.281$ & $0.32$ & 58.73 & 6.64\\
Cond SG2Im \cite{sg2im} & $14.25$ & $0.844$ & $0.081$ & $13.40$ & 11.14\\
SIMSG \cite{simsg}            & 10.812 &   0.861    &  0.065  &  6.544  & 12.07\\ 
DisPositioNet \cite{farshad2022DisPositioNet}  & 10.415 & 0.868 & 0.060 & 6.258 & 11.65\\
\methodName{} (Ours)         & \textbf{9.908}    & \textbf{0.864} & \textbf{0.064}  & \textbf{5.988}  & \textbf{13.11}\\ \hline
\end{tabular}}
\label{tab:main_VG}
\end{table}

\begin{table}[tbh]
\centering
\caption[Image reconstruction comparison between main approaches on CLEVR]{Comparison to SOTA for Image reconstruction on the CLEVR dataset}
\resizebox{\linewidth}{!}{
\begin{tabular}{l|lllll}
\hline
Method                      & MAE $\downarrow$      & SSIM $\uparrow$ & LPIPS $\downarrow$ & FID $\downarrow$ & IS $\uparrow$\\ \hline
SIMSG \cite{simsg} & 5.47 & 0.96 & 0.035 & 4.73 & 18.77 \\ 
DisPositioNet \cite{farshad2022DisPositioNet}  & $6.65$ & $0.96$ & $0.061$ & $5.61$ & 8.34\\
\methodName{} (Ours) & \textbf{4.10}    & \textbf{0.99} & \textbf{0.010}  &  \textbf{2.63} & \textbf{25.68}  \\ \hline
\end{tabular}}
\label{tab:main_CLEVR}
\end{table}

The findings are further supported by a more challenging VG dataset, which can be seen depicted in \autoref{tab:main_VG}. Both approaches significantly outperform the SIMSG~\cite{simsg} baseline in the image restoration task.

\begin{table}[tbh]
\centering
\caption[ablation study on VG]{Ablation study of different components on the VG. \textbf{P}: Progressive, \textbf{M}: Multi-head}
\resizebox{\linewidth}{!}{
\begin{tabular}{ll|llll}
\hline
M & P                     & MAE $\downarrow$      & SSIM $\uparrow$ & LPIPS $\downarrow$ & FID $\downarrow$ \\ \hline
 \multicolumn{6}{c}{VG} \\ \hline
 - & -           & 10.812 &   0.861    &  0.065  &  6.544  \\
\checkmark & -          &  9.802     & \textbf{0.866} & 0.065  &  6.302 \\
- & \checkmark         & \textbf{9.677} &   0.863    &  \textbf{0.062}  &  6.766 \\ 
\checkmark & \checkmark        & 9.908    & 0.864 & 0.064  & \textbf{5.988}  \\ \hline
 \multicolumn{6}{c}{CLEVR} \\ \hline
- & - & 5.048 &   0.982    &  0.015  &  3.575 \\
\checkmark & -      & 4.273     & 0.985 & 0.011  &  2.639  \\ 
- & \checkmark   & 4.454 &   \textbf{0.987}    &  \textbf{0.008}  &  2.757 \\ 
\checkmark & \checkmark   & \textbf{4.101}    &  0.986 & 0.010  &  \textbf{2.634} \\ \hline
\end{tabular}}
\label{tab:ablation1_VG}
\end{table}

\subsection{Qualitative Results}\label{sec:qualitative_results}
\begin{figure}[tbh]
    \centering
    \includegraphics[width=0.49\textwidth]{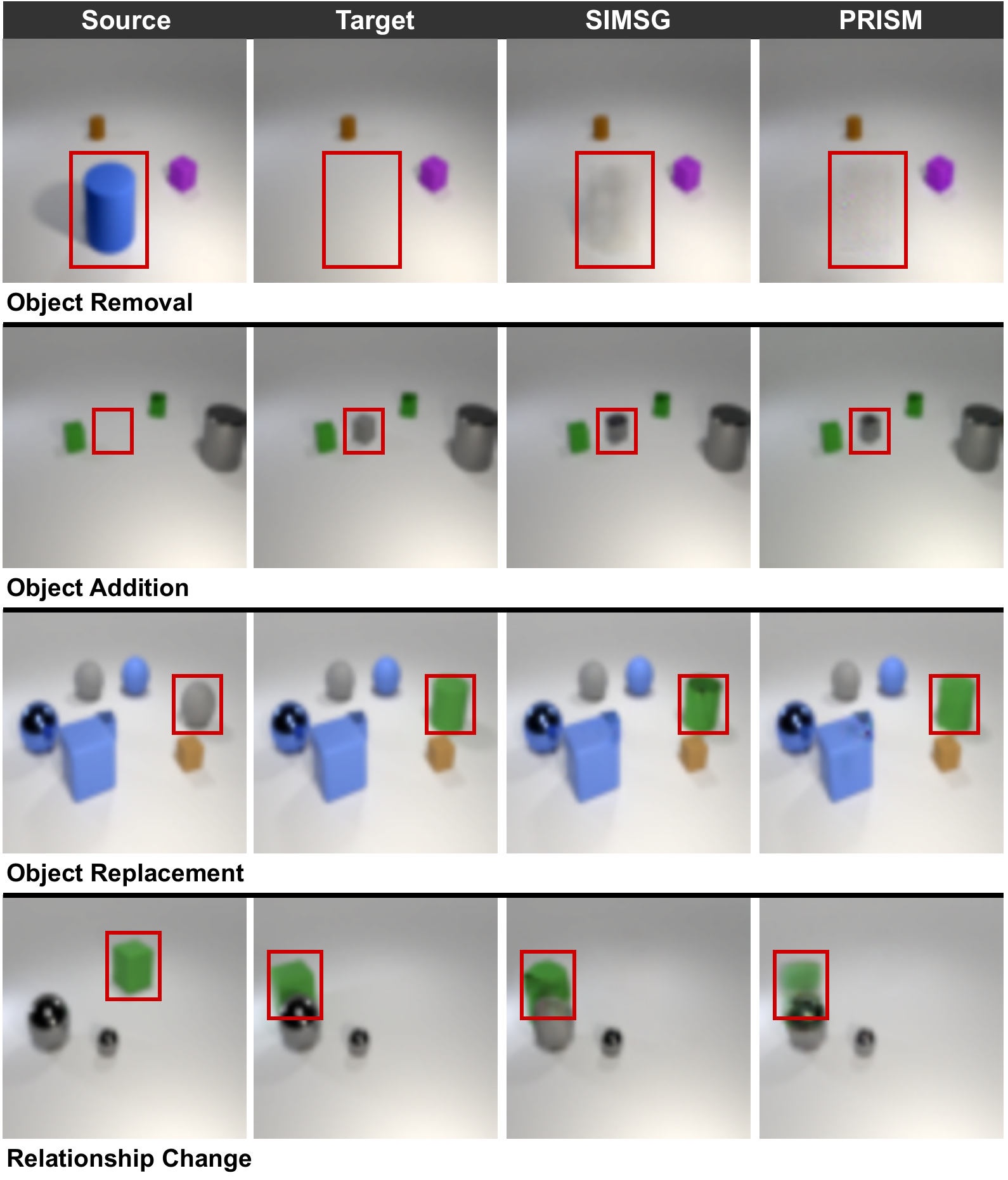}
    \caption[Image manipulation comparison of SIMSG and PRISM]{Qualitative comparison of SIMSG~\cite{simsg} and PRISM in all four image manipulation modes (removal, addition, replacement, and reposition) on the CLEVR \cite{CLEVR} dataset.}
    \label{fig:comparison_PRISM_SIMSG_CLEVR}%
\end{figure}

\paragraph{PRISM Compared to SIMSG}
In addition to generating the entire image, PRISM's two-headed approach also reconstructs a single image region containing a previously masked object. Generating a magnified detailed reconstruction of the masked object helps the network obtain information about small objects that would otherwise be lost. In \autoref{fig:comparison_PRISM_SIMSG_CLEVR} we see an example of all four image manipulation modes (removal, addition, replacement, and reposition) realized by both SIMSG and PRISM. We can see that both approaches were for the most part successful in generating the correct shape and color of the generated objects. However, SIMSG occasionally fails to capture the fine details of the object's texture. On the other hand, PRISM is significantly more successful in capturing these details. These finding are further support by a qualitative study performed on the VG dataset depicted in \autoref{fig:comparison_PRISM_SIMSG_VG}. The generated images demonstrate that PRISM exhibits superior performance in capturing the underlying structure of the image compared to SIMSG. In addition, these images show that PRISM is outperforms SIMSG in completing the desired manipulation task, further establishing its robustness.




\subsection{Ablation Study}
In this section, we analyze the different components of our methodology. First, we show the effect of the multi-head decoder and progressive unmasking on the image generation process in \autoref{tab:ablation1_VG}. Then, we show the effect of the residual connection between the two generator heads in \autoref{tab:two_head_connection}. Finally, we provide the results on the different window sizes for the second head of the generator in \autoref{tab:two_head_window_sizes}. %

\begin{table}[tbh]
  \centering
    \caption[Ablation study]{Ablation study on the effect of the residual connection between the two heads on CLEVR and VG.}%
  \label{fig:table}\label{tab:two_head_connection}
  \begin{tabular}{l|llll}
  \hline         Connection          & MAE $\downarrow$      & SSIM $\uparrow$ & LPIPS $\downarrow$ & FID $\downarrow$ \\
  \hline
  \multicolumn{5}{c}{CLEVR} \\ \hline
-            & 4.888 &   0.985    &  0.011  &  2.702  \\
\checkmark            & \textbf{4.273}     & 0.985 & 0.011  &  \textbf{2.629}  \\ \hline
  
\multicolumn{5}{c}{Visual Genome} \\ \hline
-            & 9.9 &   0.856    &  0.076  &  7.728  \\
\checkmark            & \textbf{9.677}     & \textbf{0.863} & \textbf{0.062}  &  \textbf{6.766}  \\ \hline
\end{tabular}

\end{table}

\paragraph{Residual Connection} 
As we can see in \autoref{fig:method}, the second head, which generates a detailed reconstruction of the masked input, sends the  feature representation of the object as input to the first head which generates the entire image.  Intuitively, this allows for an easier propagation of the object's visual detail information between the heads. We have done an ablation study measuring the effects of utilizing the connection which can be seen in \autoref{tab:two_head_connection}. As we can see, the results obtained from the two publicly available datasets confirm the benefits of having such a connection between the heads.

\begin{table}[tbh]
\centering
\caption[Effects of different window size values in the two-headed approach]{Effect of different window sizes ($\delta$) of the second head of the multi-head approach on CLEVR. }\label{tab:two_head_window_sizes}
\begin{tabular}{l|llll}
\hline
$\delta$                & MAE $\downarrow$      & SSIM $\uparrow$ & LPIPS $\downarrow$ & FID $\downarrow$ \\ \hline
$1.0$             & 4.068    & \textbf{0.987} & \textbf{0.010}  &  \textbf{2.506}  \\ 
$1.1$             & 4.273    & 0.986 & 0.010  &  2.629  \\ 
$1.2$             & 4.699    & 0.984 & 0.011  &  3.274  \\ 
$1.3$             & 4.413    & 0.985 & 0.016  &  2.842  \\ 
$1.5$             & \textbf{3.925}    & 0.987 & 0.011  &  2.592  \\ \hline
\end{tabular}
\end{table}

\paragraph{Delayed Alternating Optimization} 
In our investigation, we explored the effect of postponing the alternating optimization during the training phase of the two-headed architecture. Given that pre-training is a prevalent strategy in the field of computer vision, we also conducted an experiment with pre-training the first head of the two-headed system. This essentially simulated the original SIMSG training, prior to initiating the joint optimization. 
Nevertheless, we did not observe any advantages over initiating the joint optimization immediately. These outcomes resonate with the conclusions drawn by Zoph et al. \cite{pre_training_zoph2020}, who assert that joint-training possesses a greater adaptability and delivers superior results as opposed to pre-training.

\paragraph{Varying Window Sizes}
As described earlier, in the multi-head approach one head reconstructs the entire image while the second head focuses on reconstructing a single object. In this experiment, we have investigated the effects of different window sizes (i.e. $\delta=\delta\cdot \delta$) on the ability of the first head to reconstruct the entire image. As we can see from \autoref{tab:two_head_window_sizes}, having the smallest window proved to be the most beneficial. The findings we have obtained are rather intuitive, as the smallest window size is associated with the highest level of detail that the second head reconstructs, thereby offering a considerable volume of additional information. However, the outcomes derived when $\delta=1.5$ are surprisingly unexpected. The explanation for these outcomes can be traced back to an implementation nuance, whereby if the cropped window exceeds the image borders - a situation that becomes increasingly likely as the window size expands - it defaults back to $\delta=1.0$.

%% file: chapters/5_conclusion.tex
\section{Conclusion}
Our proposed multi-head semantic image manipulation method is a novel approach that utilizes a two-headed generator and progressive image inpainting techniques to improve the quality of manipulated images. Through rigorous experimentation on two public benchmarks, our method has demonstrated superior performance compared to previous works. The effectiveness of our method highlights the significant potential of these techniques for future research in the field of semantic image manipulation. Improved image manipulation techniques can have far-reaching applications in diverse fields such as image editing, computer graphics, and computer vision. Moreover, our approach can provide better quality and more efficient manipulation techniques, thereby creating new possibilities for the development of more advanced image-based applications.